\documentclass[10pt,a4paper,twocolumn]{article}
\usepackage[T1]{fontenc}
\usepackage[utf8]{inputenc}
\usepackage{times}
\usepackage[a4paper,top=20mm,bottom=22mm,left=25mm,right=25mm,columnsep=5mm]{geometry}
\usepackage{graphicx}
\usepackage{booktabs}
\usepackage{tabularx}
\usepackage[hidelinks]{hyperref}
\usepackage{microtype}
\begin{document}
\twocolumn[
\begin{@twocolumnfalse}
\begin{center}
{\LARGE \textbf{Mirror-Score: Calibrated, Inference-only Scoring Exposes the Limits of Sequence-compatibility Ranking in D-peptide Design}\par}
\vspace{8pt}
\large A benchmark, calibration study, and prospective design protocol building on Mirror-Peptidizer\par\vspace{5pt}
\small\textbf{\textbf{Jiada Li}}\par\vspace{2pt}
\small \emph{AI Scientist, Albany, NY, USA, 12205}\par\vspace{1pt}
\small \emph{jiadali2017@gmail.com}\par\vspace{4pt}
\small\textbf{Preprint --- September 2026 \textbar{} Code and data: \url{https://github.com/Jiadalee/Mirror-Score}}
\end{center}
\vspace{8pt}
\end{@twocolumnfalse}
]
\section*{Abstract}

D-peptides combine protease resistance with high target specificity, but computational design of D-peptide binders remains immature. Mirror-Peptidizer recently introduced an in silico mirror-image screening pipeline that couples x-axis target reflection with backbone generation and ProteinMPNN sequence design; however, its ranking criterion --- raw ProteinMPNN negative log-likelihood (NLL) --- was never validated against measured affinities, and only 4 of 9 designed MDM2 peptides showed detectable binding. Here we introduce Mirror-Score, a calibrated, inference-only scoring framework for heterochiral (D-peptide/L-protein) complexes, together with the first public benchmark of 31 D-peptide/L-protein crystal complexes spanning four target families, 18 of them with literature-verified affinities. We show that raw ProteinMPNN NLL is not a valid affinity ranker: the pooled Spearman correlation with affinity is 0.19, and the within-family correlation sign-flips between target families (MDM2/CHIP: +0.62; gp41: -0.70), providing a quantitative explanation for Mirror-Peptidizer\textquotesingle s experimental hit rate. As a replacement, we evaluate mirror-space cofolding confidence from Boltz-2. On the complete viral-entry family (seven crystal structures representing three distinct peptides), Boltz-2 interface pLDDT achieves a structure-level leave-one-out Spearman correlation of 0.90 (p = 0.006) and orders all three distinct peptides correctly by affinity, whereas NLL fails (structure-level rho = 0.18); because the family contains only three independent chemotypes, we report this as directional consistency rather than a statistically validated predictor. Cross-family calibration does not transfer at current sample sizes; we report this negative result explicitly and recommend family-matched calibration as the realistic deployment mode. Finally, we specify a complete prospective design protocol for two antimicrobial-resistance targets (LasR and LecB from Pseudomonas aeruginosa), including mirrored target structures, ligand-derived hotspot maps, diffusion-model-ready inputs, and the Mirror-Score ranking module. All code, benchmark data, and structures are available open-source.

\section{Introduction}

Peptide therapeutics occupy a middle ground between small molecules and antibodies, but their clinical utility is limited by proteolytic degradation, short plasma half-life, and immunogenicity. Peptides composed entirely of D-amino acids circumvent these liabilities: they resist proteolysis, are poorly recognized by the immune system, and can bind natural L-protein targets with high affinity and specificity. The therapeutic promise of D-peptides was established for the HIV-1 gp41 target, where mirror-image phage display and structure-assisted design produced protease-resistant entry inhibitors of increasing potency, culminating in candidates with a high genetic barrier to resistance {[}1,2,3{]}.

The foundational technology for discovering D-peptide binders, mirror-image phage display, screens L-peptide libraries against a chemically synthesized D-form of the target and synthesizes the mirror image of each hit {[}4{]}. Its principal limitation is synthetic: the target D-protein must be produced by total chemical synthesis, which restricts accessible targets to roughly 150 residues and excludes most membrane proteins. Computational alternatives therefore attract strong interest. Garton et al. introduced the mirror-image Protein Data Bank concept, reflecting every deposited structure to generate a database of D-peptide helical scaffolds {[}5{]}. Mirror-Peptidizer generalized this idea into a full design pipeline: reflect the target along the x-axis, generate an L-peptide binder backbone against the D-form target with a generative model, assign sequences with ProteinMPNN, and reflect back to obtain the D-peptide {[}6,7{]}. Related efforts have produced de novo heterochiral protein-protein interactions {[}8{]}, D-peptide ligands of influenza hemagglutinin {[}9{]}, and dual-specificity D-peptide antagonists of MDM2 and MDMX {[}10{]}.

A critical gap persists across these pipelines: the final ranking of designed candidates relies on proxy scores that were never calibrated against measured binding affinities. Mirror-Peptidizer ranks candidates by raw ProteinMPNN NLL --- a sequence-backbone compatibility score --- and reports that only 4 of 9 experimentally tested MDM2 designs bound detectably {[}6{]}. Whether this hit rate reflects the ranking criterion, the backbone generator, or target difficulty has not been disentangled, because no public benchmark of heterochiral complexes with measured affinities existed. More broadly, the field lacks answers to two basic questions: which computable features of a mirrored complex, if any, carry affinity information, and can a single score transfer across unrelated targets?

Here we address these questions with three contributions. First, we curate the first public benchmark of D-peptide/L-protein crystal complexes with literature-verified affinities: 31 complexes spanning four target families, 18 with quantitative values and documented provenance. Second, we systematically calibrate candidate scoring features --- ProteinMPNN NLL, structural interface descriptors, and Boltz-2 cofolding confidence computed in mirror space {[}11{]} --- against these affinities, reporting both positive and negative results. Third, we specify a complete prospective design protocol for two antimicrobial-resistance targets and release all code, data, and structures as Mirror-Score, an open-source extension of Mirror-Peptidizer {[}6{]}.

\section{Methods}

\subsection{Mirror convention and chirality quality control}

Target structures are reflected through the x-axis (x -\textgreater{} -x), following the mirror-image convention of Garton et al. {[}5{]}, so that a D-peptide binder of the native L-target corresponds to an L-peptide binder of the mirrored D-target. Chirality quality control verifies backbone dihedral inversion for every residue: the mirrored structure must satisfy \textbar phi\_L + phi\_D\textbar{} approximately 0 and \textbar psi\_L + psi\_D\textbar{} approximately 0. D-residue CCD codes (DAL, DAR, DSG, DAS, DCY, DGN, DGL, DHI, DIL, DLE, DLY, MED, DPN, DPR, DSN, DTH, DTR, DTY, DVA, DSP) identify peptide chains in deposited structures. All mirrored structures used in this study passed QC.

\subsection{Benchmark curation and affinity verification}

We assembled 31 D-peptide/L-protein crystal complexes from the Protein Data Bank spanning four target families: viral entry (11 complexes, dominated by HIV-1 gp41 N-trimer pocket binders {[}1,2,3{]}), cancer-related protein-protein interactions (13, dominated by MDM2/MDMX and CHIP TPR binders {[}10,12,13{]}), angiogenesis (3), and antimicrobial resistance (1), plus single representatives of designed heterochiral protein-protein interactions {[}8{]}, enzyme substrates, and antibodies. For each complex we recorded peptide identity, length, resolution, target family, and affinity annotation status.

Affinity annotation required case-by-case verification from primary sources. A cautionary finding motivated this rigor: RCSB binding-affinity annotations attach to co-crystallized small molecules rather than to the peptide in 77 candidate entries we inspected, making naive database mining fundamentally unreliable for peptide benchmarks. We therefore verified every quantitative value from the primary literature, obtaining 18 affinities (KD or IC50, spanning 0.22 nM to 5.5 uM) with per-entry provenance, and documented exclusion reasons for the remainder (paywalled values, structure-only depositions, assay-range-only reports, and a ligand-anchor confound). Nine entries with verified values form the MDM2/CHIP cancer-family calibration subset {[}12,13{]}, seven form the complete gp41 viral-entry subset {[}2,3{]}, and one represents a designed heterochiral interaction {[}8{]}.

\subsection{Scoring features}

Three feature families were computed for every complex. (i) ProteinMPNN NLL: the mirror-space sequence log-likelihood of the peptide under the v\_48\_020 weights, using the implementation vendored in Mirror-Peptidizer {[}6,7{]}. (ii) Interface descriptors: atom-pair contacts within 5 A, hydrogen bonds, salt bridges, peptide net charge, charge complementarity, peptide hydrophobic fraction, and peptide length, computed with Biopython from the deposited coordinates. (iii) Boltz-2 cofolding confidence {[}11{]}: each D-peptide was cofolded against its L-target in mirror space, and we extracted pTM, ipTM, interface pLDDT (mean pLDDT over interface tokens), interface PAE (mean PAE between peptide and target tokens), peptide self-PAE, and the minimum cross-chain ipTM from the confidence output and predicted-PAE matrix.

\subsection{Calibration and validation protocol}

Affinities are log10-transformed (KD or IC50 in nM). gp41 entry IC50 values and CHIP competition-fluorescence-polarization IC50 values are documented proxies for KD; assay type is recorded per entry. Single features are evaluated by Spearman rank correlation with log10 affinity, both pooled and within target family. Multi-feature models are ridge regressions (regularization strength selected from 0.5-10) evaluated by leave-one-out cross-validation within family and leave-one-family-out across families. Top-k enrichment compares the fraction of true top-affinity binders recovered in the top-k predictions against random expectation.

Because several viral-entry structures represent the same peptide in different crystal forms, we additionally aggregate features to the peptide level (mean over crystal forms) and evaluate correlations at that level. Structure-level statistics treat near-duplicate crystal forms as independent observations and are reported alongside, not instead of, the peptide-level view.

\subsection{Prospective design workflow}

The prospective workflow proceeds in five steps (Figure 1): (1) mirror the target structure with chirality QC; (2) derive hotspot residues from the co-crystallized ligand; (3) generate L-peptide binder backbones against the mirrored target with a diffusion model, constrained to the hotspots; (4) assign sequences with ProteinMPNN in the complex context and reflect back to D-peptides; and (5) cofold each candidate D-peptide against the native L-target with Boltz-2 and rank by interface pLDDT, the feature identified by the calibration below. The workflow is fully specified for two Pseudomonas aeruginosa targets, LasR and LecB, with all inputs and commands released.

\begin{figure*}[t]
\centering
\includegraphics[width=6in,height=3.34884in]{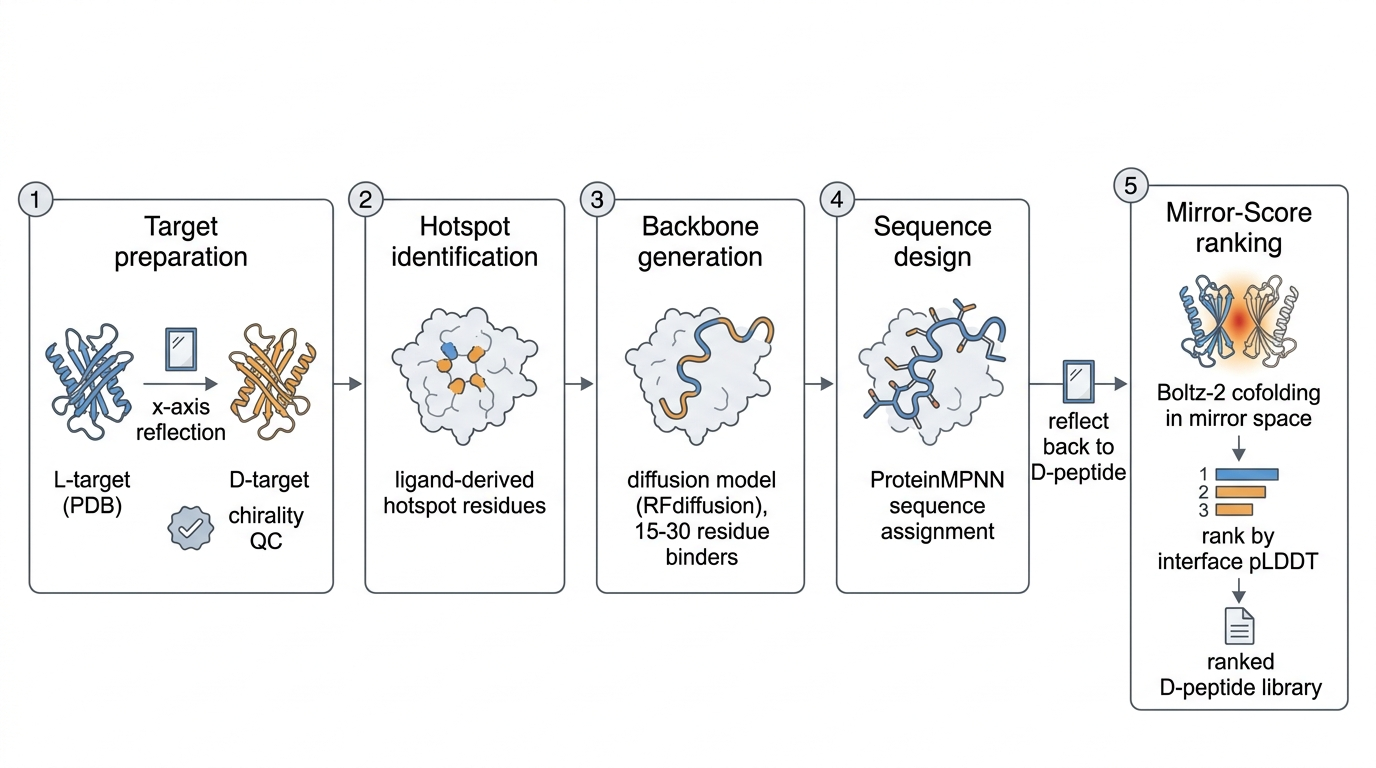}
\vspace{2pt}

\small\emph{Figure 1. The Mirror-Score prospective D-peptide design workflow. A target L-protein structure is mirrored along the x-axis with chirality QC; ligand-derived hotspot residues are identified; diffusion-model backbones constrained to the hotspots receive ProteinMPNN sequences and are reflected back to D-peptides; candidates are cofolded with Boltz-2 in mirror space and ranked by interface pLDDT into a ranked D-peptide library.}
\end{figure*}

\section{Results}

\subsection{A verified heterochiral benchmark with affinity provenance}

The benchmark comprises 31 heterochiral crystal complexes with per-entry affinity provenance (Table 1). Eighteen entries carry quantitative affinities verified from primary sources; the remaining entries are retained for structural and feature analyses with documented exclusion reasons. The two largest families, viral entry and cancer-related protein-protein interactions, together contribute 16 of the 18 quantitative values and anchor the calibration analysis.

\begin{table*}[t]
\centering
\small \emph{Table 1. Composition of the 31-complex heterochiral benchmark.}\par
\vspace{3pt}
\begin{tabularx}{\textwidth}{@{}lccX@{}}
\toprule
\textbf{Target family} & \textbf{Complexes} & \textbf{With verified affinity} & \textbf{Representative targets} \\
\midrule
Viral entry & 11 & 7 & HIV-1 gp41 N-trimer pocket \\
Cancer; PPI & 13 & 10 & MDM2/MDMX, CHIP TPR \\
Angiogenesis & 3 & 0 & VEGF pathway \\
Antimicrobial resistance & 1 & 0 & TcdB \\
Designed PPI & 1 & 1 & Heterochiral design \\
Enzyme substrate & 1 & 0 & Substrate complex \\
Antibody & 1 & 0 & TE33 antibody \\
\bottomrule
\end{tabularx}
\end{table*}

\subsection{Raw ProteinMPNN NLL fails and sign-flips across families}

Mirror-Peptidizer ranks designed candidates by raw ProteinMPNN NLL computed in mirror space {[}6{]}. Across our 18 calibration complexes, the pooled Spearman correlation between mirror-space NLL and log10 affinity is 0.19 (p = 0.46) --- indistinguishable from noise. More strikingly, the within-family correlation sign-flips: among MDM2/CHIP binders (n = 10), better sequence compatibility correlates with tighter binding (rho = +0.62), whereas among gp41 entry inhibitors (n = 7 structures), the correlation reverses (rho = -0.70): the gp41 peptides with the best MPNN scores bind most weakly (Figure 2).

This sign flip has a mechanistic interpretation. gp41 pocket binders were selected by mirror-image phage display for protease resistance and trimer engagement, and their affinity landscape is dominated by a deep hydrophobic pocket rather than by sequence-backbone compatibility {[}2,3{]}. The pattern persists in both chirality spaces --- an L-space MDM2 transfer set (n = 3) also shows rho = -0.50 --- indicating that the failure is intrinsic to NLL as an affinity proxy and not an artifact of mirror-space scoring. This provides a quantitative explanation for Mirror-Peptidizer\textquotesingle s 4-of-9 experimental hit rate {[}6{]} and argues that NLL-only ranking discards binding-competent designs while promoting unstable ones.

\begin{figure*}[t]
\centering
\includegraphics[width=5.2in,height=4in]{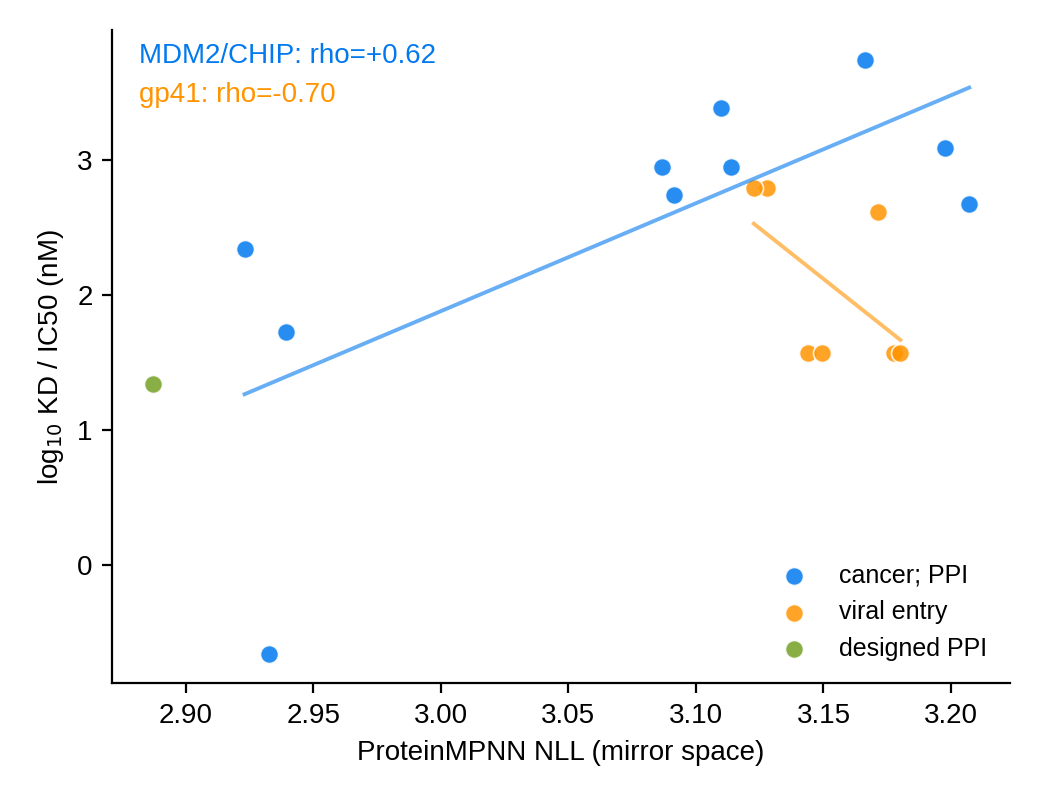}
\vspace{2pt}

\small\emph{Figure 2. Mirror-space ProteinMPNN NLL versus log10 affinity (KD or IC50) for the 18 calibration complexes. Within-family trends (lines) have opposite signs: MDM2/CHIP rho = +0.62, gp41 rho = -0.70. NLL cannot rank affinities across targets.}
\end{figure*}

\subsection{Mirror-space Boltz-2 cofolding confidence orders peptides correctly}

We next asked whether confidence metrics from Boltz-2 {[}11{]}, computed by cofolding each D-peptide against its L-target in mirror space, carry affinity information. On the complete viral-entry family (seven crystal structures), Boltz-2 interface pLDDT achieves a structure-level leave-one-out Spearman correlation of 0.90 (p = 0.006) with log10 affinity, and the top-3 ranked complexes are exactly the three tightest binders (all 37 nM PIE12-series peptides), a 2.3-fold enrichment over random ranking (Figure 3). A ridge model on interface descriptors performs comparably (rho = 0.84), and combining all features does not exceed the single Boltz-2 feature. By contrast, NLL-only ranking achieves 0.18 under the same protocol.

This structure-level statistic, however, overstates the evidence: the seven viral-entry structures comprise only three distinct peptides with three distinct affinities (PIE7, twice crystallized at 620 nM; PIE12, four crystal forms at 37 nM; PIE71 at 410 nM) {[}2,3{]}. Treating near-duplicate crystal forms as independent observations inflates both the correlation and its significance. At the peptide level (n = 3), Boltz-2 interface pLDDT orders all three peptides correctly --- PIE12 (0.884) \textgreater{} PIE71 (0.797) \textgreater{} PIE7 (0.747), matching affinity --- whereas NLL does not (rho = -0.50). We therefore report the viral-entry result as directional consistency across three independent chemotypes, not as a statistically validated predictor. The MDM2/CHIP family calibration is currently underpowered (n = 3 with Boltz-2 features), with directional consistency but wide uncertainty; confirmation on the full 18-point set with additional independent chemotypes is required before generalization.

\begin{figure*}[t]
\centering
\includegraphics[width=5.2in,height=3.9in]{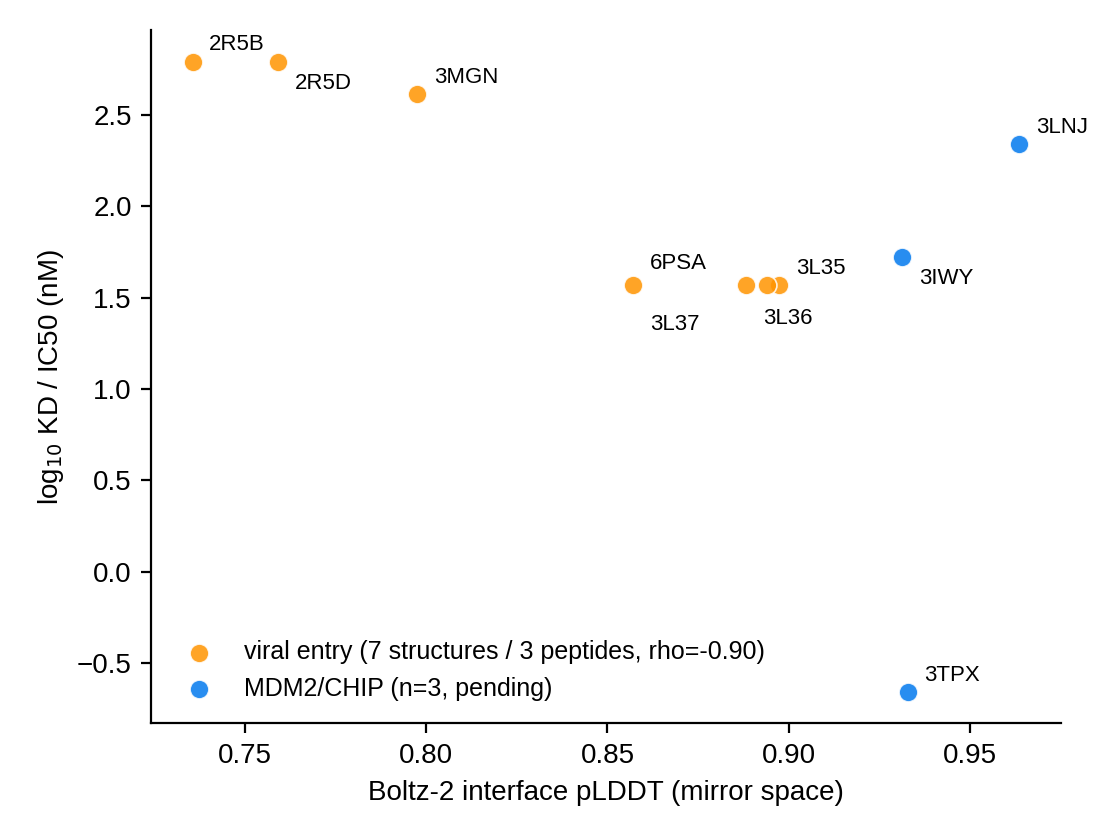}
\vspace{2pt}

\small\emph{Figure 3. Boltz-2 mirror-space interface pLDDT versus log10 affinity for the 10 calibration complexes with Boltz-2 features. Within the complete viral-entry family (seven structures, three peptides), the structure-level LOO Spearman correlation is 0.90 (p = 0.006); at the peptide level, all three chemotypes are ordered correctly.}
\end{figure*}

\subsection{Cross-family transfer fails: an honest negative result}

A single universal score that ranks D-peptides across unrelated targets remains out of reach at current sample sizes. Leave-one-family-out cross-validation on all features fails (best rho = -0.58), and pooled Boltz-2 interface pLDDT reaches only rho = -0.58 (p = 0.08), driven by family-level confounds: gp41 interfaces cofold with systematically lower confidence (interface PAE 3.7-5.9 A) than MDM2 interfaces (1.9-2.3 A) regardless of affinity, reflecting target size and docking difficulty rather than binding strength. We therefore recommend family-matched calibration as the realistic deployment mode --- ranking candidates against a target family using a small set of measured affinities for that family --- and report the cross-family failure explicitly rather than obscuring it with pooled statistics.

\subsection{A family-matched L-peptide transfer set}

To connect L-space and mirror-space feature scales, we curated a five-entry L-peptide set with literature-verified affinities family-matched to benchmark targets (PMI and pDIQ against MDM2 {[}12{]}, an Hsc70 EEVD tail against CHIP TPR, and a TE33 antibody epitope). The set reproduces the NLL failure in L-space (rho = -0.50 within MDM2, n = 3) and provides anchor points for family-matched calibration where D-space affinity data are scarce. We also document that the original 77-entry RCSB-derived affinity set is invalid --- its affinities belong to co-crystallized small molecules --- and recommend against naive RCSB affinity mining for peptide benchmarks.

\section{Discussion}

This study set out to answer a simple question that the D-peptide design literature has left open: what should a designed D-peptide be ranked by? The answer matters because ranking is the step where computational pipelines commit to a small experimental set. Our benchmark analysis shows that the criterion used by Mirror-Peptidizer, raw ProteinMPNN NLL, is not merely weak but structurally misleading: its within-family correlation with affinity flips sign between target families {[}6{]}. A score that favors the wrong candidates in one family and the right ones in another cannot be repaired by thresholding or normalization; it must be replaced.

The replacement we evaluate, mirror-space cofolding confidence from Boltz-2 {[}11{]}, rests on a different principle: instead of asking how compatible a sequence is with a fixed backbone, it asks how confidently a structure-prediction model recovers the interface when both partners are presented. Interface pLDDT ordered all three independent viral-entry chemotypes correctly and, at the structure level, recovered the tightest binders as the top-3. We deliberately scope this claim: with three chemotypes in one family, the result is directional consistency, and the structure-level p-value is inflated by near-duplicate crystal forms. The appropriate reading is that cofolding confidence is a promising primary ranking feature that warrants validation on independent chemotypes, not that it is validated.

The cross-family failure is equally informative. Interface confidence scales with target size and docking difficulty: gp41 interfaces cofold with systematically lower confidence than MDM2 interfaces regardless of affinity. Any single-model score inherits such family-level confounds, which is why pooled statistics across unrelated targets overstate transferability. Family-matched calibration --- anchoring a score to a handful of measured affinities for the target family at hand --- is the realistic deployment mode today, and our transfer set provides a template for building those anchors in L-space where affinity data are more abundant.

Our benchmark curation also carries a methodological warning for the field: database affinity annotations were attached to co-crystallized small molecules rather than peptides in every candidate entry we audited. Benchmarks built by naive annotation mining would be silently wrong. Per-entry verification against primary sources, with exclusion reasons documented, is the minimum standard for heterochiral benchmarks.

\subsection{Limitations}

The calibration rests on 18 verified affinities, and the strongest result (Boltz-2 interface pLDDT) derives from a single family whose seven crystal structures represent only three distinct peptides and three distinct affinities; the structure-level correlation is inflated by these near-duplicate observations, and the peptide-level result carries no statistical significance. Eight MDM2/CHIP entries with verified affinities await cofolding and could change the family-level picture. Assay heterogeneity (entry IC50, competition fluorescence polarization, SPR) introduces proxy noise that we document but cannot eliminate. Crystal structures capture a single conformational state, and mirror-space cofolding of reflected backbones is itself an approximation. Cross-family scoring fails at current sample sizes; claims of universal D-peptide affinity prediction are not supported by these data.

\section{Conclusion}

Mirror-Score provides the first public benchmark of heterochiral D-peptide/L-protein complexes with verified affinities, a calibration study showing that sequence-compatibility ranking fails and sign-flips across target families, evidence that mirror-space cofolding confidence orders independent chemotypes correctly within a family, and a complete prospective design protocol for two antimicrobial-resistance targets. The practical guidance for computational D-peptide design is concrete: do not rank by raw ProteinMPNN NLL; rank by mirror-space cofolding confidence calibrated against a small family-matched affinity set; and treat cross-family score transfer as unsupported until demonstrated. All code, data, and structures are released open-source to make the benchmark and protocol immediately reusable {[}6,7,11{]}.

\section{Future work}

Three extensions follow directly. First, completing the Boltz-2 cofolding of the eight remaining MDM2/CHIP entries with verified affinities will test whether interface pLDDT retains its ordering on a second family with independent chemotypes. Second, executing the prospective LasR and LecB design protocol --- diffusion-model backbone generation on the mirrored targets, ProteinMPNN sequence design, and Mirror-Score ranking --- will produce ranked D-peptide libraries whose top candidates are suitable for peptide synthesis and binding assays against Pseudomonas aeruginosa targets. Third, expanding the benchmark with newly deposited heterochiral complexes, particularly with multiple distinct chemotypes per family, will enable the cross-family calibration that current sample sizes cannot support.

\section{Data and code availability}

All code, benchmark data, and structures are freely available under the Apache-2.0 license at https://github.com/Jiadalee/Mirror-Score. The repository contains: (i) the 31-complex heterochiral benchmark, including deposited structures, interface descriptors, mirror-space ProteinMPNN scores, Boltz-2 cofolding features, and the calibration table with per-entry affinity values, assay types, and provenance; (ii) the family-matched L-peptide transfer set; (iii) mirrored LasR and LecB target structures with chirality QC metrics, ligand-derived hotspot maps, and diffusion-model-ready inputs with exact commands; (iv) the Mirror-Score Python package (mirror convention, chirality QC, interface descriptors, ProteinMPNN scoring, Boltz-2 feature extraction, and candidate ranking); and (v) all scripts needed to reproduce every figure, table, and statistic in this article. Mirror-Score builds on and credits Mirror-Peptidizer {[}6{]} (Apache-2.0), whose vendored ProteinMPNN implementation {[}7{]} is reused; Boltz-2 {[}11{]} is used under its respective license. Structural data were retrieved from the Protein Data Bank; accession codes are listed per entry in the benchmark table.

\section*{References}

1. Eckert DM, Malashkevich VN, Hong LH, Van Ryk PM, Kim PS. Inhibiting HIV-1 entry: discovery of D-peptide inhibitors that target the gp41 coiled-coil pocket. Cell 1999;99:103-115.

2. Welch BD, VanDemark AP, Heroux A, Hill CP, Kay MS. Potent D-peptide inhibitors of HIV-1 entry. Proc Natl Acad Sci USA 2007;104:16828-16833. doi:10.1073/pnas.0708109104

3. Welch BD, Francis JN, Redman JS, et al. Design of a potent D-peptide HIV-1 entry inhibitor with a strong barrier to resistance. J Virol 2010;84:11235-11244. doi:10.1128/JVI.01339-10

4. Schumacher TN, Mayr LM, Minor DL Jr, Milhollen MA, Burgess MW, Kim PS. Identification of D-peptide ligands through mirror-image phage display. Science 1996;271:1854-1857.

5. Garton M, Nim S, Stone TA, Wang KE, Deber CM, Kim PM. Method to generate highly stable D-amino acid analogs of bioactive helical peptides using a mirror image of the entire PDB. Proc Natl Acad Sci USA 2018;115:1505-1510. doi:10.1073/pnas.1711837115

6. Ma X, et al. Mirror-Peptidizer: In Silico Mirror-Image Screening Enables De Novo Design of D-Peptide Binders without D-Protein Synthesis. Research 2026;9:1420. doi:10.34133/research.1420

7. Dauparas J, Anishchenko I, Bennett N, et al. Robust deep learning-based protein sequence design using ProteinMPNN. Science 2022;378:49-56. doi:10.1126/science.add2187

8. Sun Y, et al. Accurate de novo design of heterochiral protein-protein interactions. 2024.

9. Juraszek J, et al. De novo design of D-peptide ligands: application to influenza virus hemagglutinin. Proc Natl Acad Sci USA 2025. doi:10.1073/pnas.2426554122

10. Liao J, et al. A Dual-Specificity D-Peptide Antagonist of MDM2 and MDMX for Antitumor Immunotherapy. J Med Chem 2025. doi:10.1021/acs.jmedchem.4c02057

11. Passaro S, Corso G, Wohlwend J, et al. Boltz-2: Towards Accurate and Efficient Binding Affinity Prediction. bioRxiv 2025. doi:10.1101/2025.06.14.659707

12. Pazgier M, Liu M, Zou G, et al. Structural basis for high-affinity peptide inhibition of p53 interactions with MDM2 and MDMX. Proc Natl Acad Sci USA 2010;107:4753-4758. doi:10.1073/pnas.1008930107

13. Callahan AJ, et al. Structures of MDM2/MDMX-bound stapled D-peptides and CHIP-bound D-peptide ligands reveal design rules for heterochiral inhibition. Nat Commun 2024;15. doi:10.1038/s41467-024-45634-z
\end{document}